\documentclass{IEEEtran}
\usepackage{etoolbox}
\makeatletter
\patchcmd{\@makecaption}
  {\scshape}
  {}
  {}
  {}
\patchcmd{\@makecaption}
  {\\}
  {.\ }
  {}
  {}
\patchcmd{\@makecaption}
  {\centering}
  {}
  {}
  {}
\makeatother

\makeatletter
\def\thickhline{%
  \noalign{\ifnum0=`}\fi\hrule \@height \thickarrayrulewidth \futurelet
   \reserved@a\@xthickhline}
\def\@xthickhline{\ifx\reserved@a\thickhline
               \vskip\doublerulesep
               \vskip-\thickarrayrulewidth
             \fi
      \ifnum0=`{\fi}}
\makeatother

\newlength{\thickarrayrulewidth}
\usepackage{ifpdf}
    \ifpdf 
      \usepackage{graphicx}  
      \usepackage{epstopdf}
    \else
      \usepackage[dvips]{graphicx}  
    \fi 
    
\usepackage{graphicx}
\usepackage[subrefformat=parens,labelformat=parens,caption=false,font=footnotesize]{subfig}
\usepackage{cite}
\usepackage{tabulary}
\usepackage{amssymb}
\usepackage{float}
\usepackage{epstopdf}
\usepackage{color}
\usepackage[bookmarks=false]{hyperref}
\usepackage{balance}
\usepackage{cite}
\usepackage[cmex10]{amsmath}
\usepackage{url}
\usepackage{bm}
\usepackage{enumitem}
\setitemize{leftmargin=*}
\graphicspath{{./images/}}
\usepackage{flushend} 

\IEEEoverridecommandlockouts 
\begin{document}
\captionsetup[subfloat]{labelformat=empty}

\bstctlcite{IEEEexample:BSTcontrol}

\title{\LARGE \bf
Learning Perceptual Locomotion on Uneven Terrains using Sparse Visual Observations}
\author{Fernando Acero, Kai Yuan, Zhibin Li 
\thanks{This work was supported by the EPSRC CDT in Robotics and Autonomous Systems. Authors are with the School of Informatics, University of Edinburgh, and Department of Computer Science, University College London, United Kingdom. 
Email: {\tt\small \{name.surname\}@ed.ac.uk}
}
}


\maketitle

\begin{abstract}

To proactively navigate and traverse various terrains, active use of visual perception becomes indispensable. We aim to investigate the feasibility and performance of using sparse visual observations to achieve perceptual locomotion over a range of common terrains (steps, ramps, gaps, and stairs) in human-centered environments. We formulate a selection of sparse visual inputs suitable for locomotion over the terrains of interest, and propose a learning framework to integrate exteroceptive and proprioceptive states. We specifically design the state observations and a training curriculum to learn feedback control policies effectively over a range of different terrains. We extensively validate and benchmark the learned policy in various tasks: omnidirectional walking on flat ground, and forward locomotion over various obstacles, showing high success rate of traversability. Furthermore, we study exteroceptive ablations and evaluate policy generalization by adding various levels of noise and testing on new unseen terrains. We demonstrate the capabilities of autonomous perceptual locomotion that can be achieved by \textit{only} using sparse visual observations from direct depth measurements, which are easily available from a Lidar or RGB-D sensor, showing robust ascent and descent over high stairs of 20 cm height, i.e., 50\% leg length, and robustness against noise and unseen terrains. 
\end{abstract}


\IEEEpeerreviewmaketitle
\section{Introduction}

\begin{figure}[t]
\centering
\includegraphics[width=86mm]{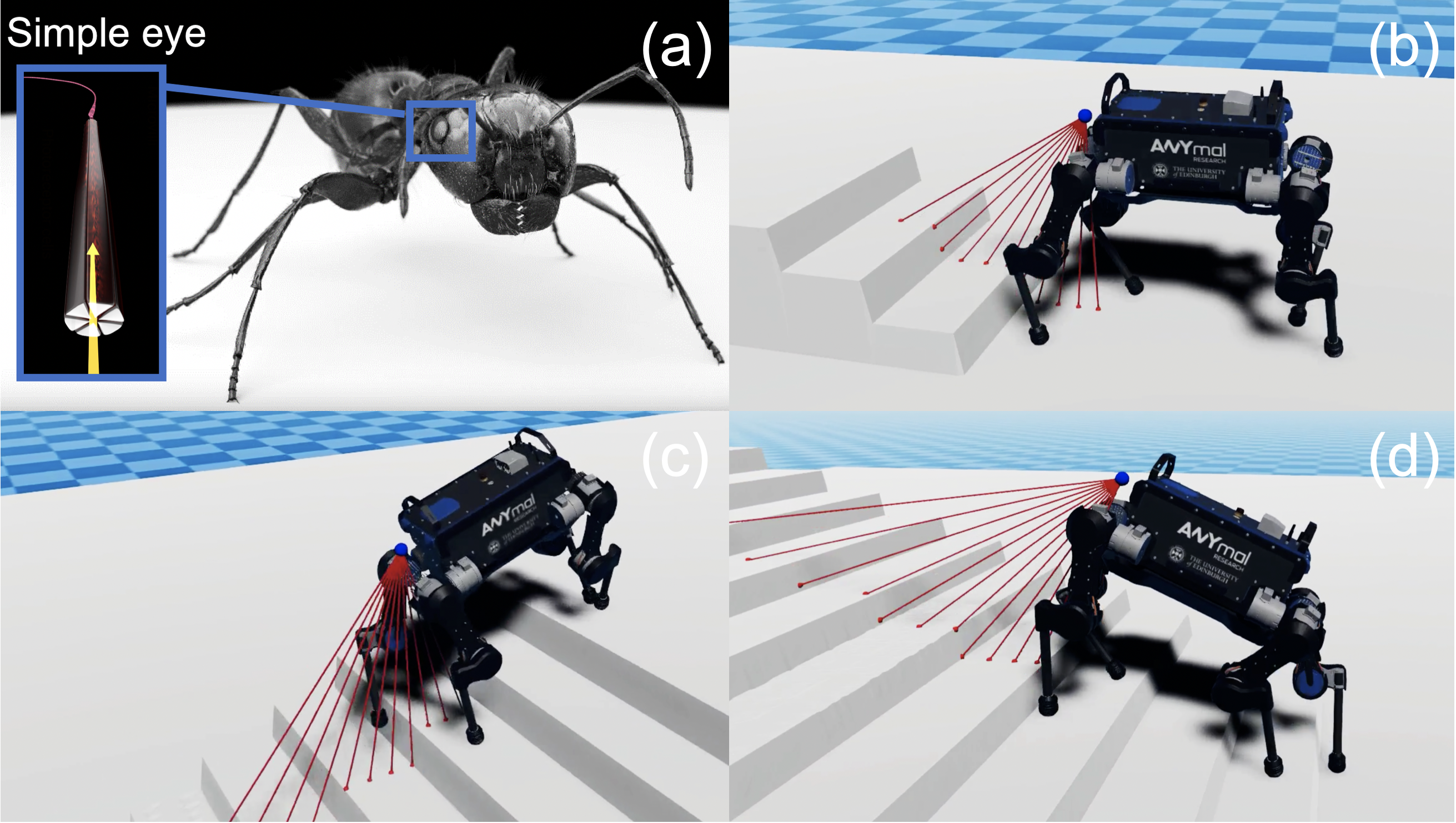} 
\caption{Depth-sensing-driven perceptual locomotion and its bio-inspiration: (a) compound eyes consist of simple photo-receptor units, (b) robot approaching stairs, (c) descending stairs, (d) climbing stairs -- all using a small set of simple visual inputs.
}
\label{fig:insecteye}
\vspace{-4mm}
\end{figure}

Evidence from nature suggest that even simple vision systems can provide meaningful information to enable perceptual motor control \cite{giurfa1997insect, warrant2017remarkable}. For instance, compound eyes in most insects consist of multiple but simple eyes geometrically arranged to provide visual information over a wide field of view and with low resolution (Figure \ref{fig:insecteye}(a)). Yet insects are able to exhibit various locomotion and flight behaviours using such simple forms of visual feedback. This work is motivated to investigate a particular research question: can perceptual legged locomotion be learned using \textit{only} sparse visual observations?

Traditionally, control and perception are approached separately and then integrated in a modular manner, which can be computationally expensive to run in real-time and requires powerful on-board computers. 
In contrast, a learning-based approach using neural networks offers an alternative to bridge sensing and closed-loop control by feeding perceptual data directly into the neural network based control policy. In this way, we can train a policy that has access to both proprioceptive and exteroceptive data within the feedback loop, and thus is an integral manner to learn environment-aware behaviours. Here, we use the term \textit{perceptual locomotion} for such capabilities.

The capability of blind walking, i.e., by using proprioceptive feedback only, has achieved remarkable performance using either mixed-integer optimization \cite{MIT_Opti, RPC} and model-based control \cite{MPC_Guiyang, NMPC, Angelini_MPC, MIT}, or machine learning which heavily utilizes physics simulations to train control policies by reinforcement learning (RL) \cite{hwangbo2019learning, ETH, DRL_Haarnoja, DRL_Xie, MELA, RMA} or imitation learning \cite{ETH_Imitation, SonyImitation, peng2020learning}. 
While optimization-based controllers are very good at guaranteeing stability, they are limited in scalability to unmodeled settings in the real world, due to expensive optimization \cite{MIT}. Noticeably, optimization-based approaches can be capable of traversing the terrains studied in this work, but they would require detailed mapping of the environment and require substantial or offline computation of optimization-based planning. 

As an attractive alternative, learning-based controllers can leverage large amounts of experience gathered in simulation to produce control policies that can exhibit robust behaviours \cite{ETH, MELA, RMA} without online computational overhead as in optimization-based control. Learning-based controllers have demonstrated certain degree of environment awareness, although these approaches typically consist of blind locomotion and passive reflexes \cite{CPG_Victor_Claudio, ETH}. Such inherent limitation is that reflexes are only triggered after interactions with the environment take place, e.g., a fore leg first collides against a step before a reaction occurs. Hence, these approaches do not form active control and have limited environmental awareness. In contrast, the proposed perceptual locomotion here can perceive the presence of a step through exteroceptive inputs, and actively modulate the gait online and proactively traverse difficult terrains. 

There are existing works that have integrated visual feedback into controllers in some distinct way to our approach, typically requiring pre-processing or ground-truth data and adding online computational complexity such as by building maps or extracting encodings \cite{miki2022learning}, instead of providing the locomotion policy with exteroceptive observations directly. Camera feeds have been used as input to a neural network trained to predict safe footholds \cite{stereoVision2, siravuru2017deep}. The importance of Lidar sensors to produce environment-aware controllers has been highlighted \cite{sensorFusionVision}. Hybrid methods can combine RL and optimization, and use terrain height maps to produce adaptive locomotion behaviours \cite{ox, tsounis2020deepgait}. The use of height maps has been explored to develop hierarchical controllers for bipedal locomotion as well \cite{deeploco}. Outside of robotics applications, learning-based approaches have also been explored to map sparse visual perception of ground surfaces to whole-body motion synthesis of animation characters \cite{holden2017phase}.  

\vspace{-0.5mm}

\subsection{Motivation and Contribution}
Inspired by the vast capabilities that simple visual feedback as that of insects enables in nature \cite{giurfa1997insect, warrant2017remarkable}, we use a robot as a platform to study the capabilities that sparse visual observations can endow to legged locomotion. This work aims to answer the following research questions: (1) What is the basic type of visual representation to achieve perceptual locomotion on uneven human-centered terrains? (2) For the traversal of common steps and stairs in human-centered environments, are sparse visual observations sufficient to render visual abstraction? (3) What is the learning architecture that can integrate such visual abstraction easily and effectively? Especially being compatible with many existing \textit{blind locomotion} schemes?

As a long-term goal to deploy useful applications in unstructured environments, we envision that the next stage of legged locomotion research is to investigate effective and efficient exploitation of visual information -- to achieve robust and agile perceptual locomotion. 
Hereby, this work investigates a learning-based approach to directly integrate visual perception, by using sparse observations for effective terrain-aware perceptual locomotion. To do so, we use a ANYmal B robot and the RaiSim physics engine \cite{raisim} for simulations, due to its good accuracy and efficiency for multi-body dynamics simulation. Our method is not specifically developed for one robot and should be applicable to other quadrupeds with joint position or impedance control. 

The contributions of this work are summarised as follows:
\begin{enumerate}
\item A study with positive results to show the feasibility of using only sparse visual inputs of surfaces ahead of the robot, and training of a perceptual locomotion policy that can traverse uneven human-centered terrains with high success rates even on unseen terrains; 
\item A design guide to estimate the set of depth observations which are sufficient for abstract representation of steppable terrain irregularities for perceptual locomotion;
\item A novel learning curriculum that enables the training of a single policy that succeeds on multiple tasks, including omnidirectional walking on flat grounds, and walking over steps, ramps and stairs, as well as unseen terrains with ditches, barriers, and alternating stairs, while demonstrating robustness to exteroceptive noise and ablations.
\end{enumerate}

The remainder of the paper is organized as follows. In Section~\ref{sec:learning_perceptual_locomotion}, we provide our formulation of the chosen visual perception, a discussion of the learning-based controller architecture, a comprehensive description of the RL paradigm and our custom training curriculum. In Section~\ref{sec:results}, we provide a summary of our training process, a detailed analysis of locomotion over high stairs, and present the results of our extensive tests. Finally, we conclude in Section~\ref{sec:conclusion}.

\vspace{-1mm}
\section{Learning Perceptual Locomotion}

\label{sec:learning_perceptual_locomotion}
Training perceptual locomotion via RL needs several considerations. First, the main elements in blind locomotion policies shall be relevant too, e.g., choice of proprioceptive state or reward terms. Second, the chosen representation for perceptual inputs, its compactness, and capability to capture environmental features is critical for successful policies. Yet this is relevant, as exteroceptive information often comes with much higher dimensions than proprioceptive data of the robot, e.g., images contain more pixels than the number of actuators or encoders on a robot.

However, as the observation space of the policy increases, training becomes increasingly more expensive due to the curse of dimensionality \cite{bellman1966dynamic}. Hence, it is desirable to choose a representation that is relatively small to enable training of the policy, and is also sufficient to capture features from the environment that are relevant to perceptual locomotion, e.g., to stepping over stairs. In this sense, we propose a set of visual inputs that is designed to capture terrain with sufficient resolution as elaborated in Section \ref{sec:principle:aaa}, while also being small enough not to impact training time excessively. Lastly, we design a specific training curriculum to enable a single policy to robustly locomote over various terrains. 

Several key assumptions are: terrains -- those in human-centered environments, mainly consisting of flat ground, steps, ramps, stairs, barriers and ditches to be traversed in a forward direction, as commonly needed in industrial applications; robots -- medium size quadrupeds that can physically traserve in the aforementioned surfaces; visual feedback -- sparse exteroceptive perception which can be obtained with matching specifications of common commercial cameras. More details are in the following sub-sections.

\vspace{-0.5mm}
\subsection{Formulation of Sparse Visual Perception}
\label{sec:principle:aaa}

\begin{figure}[t]
\centering
\includegraphics[width=80mm]{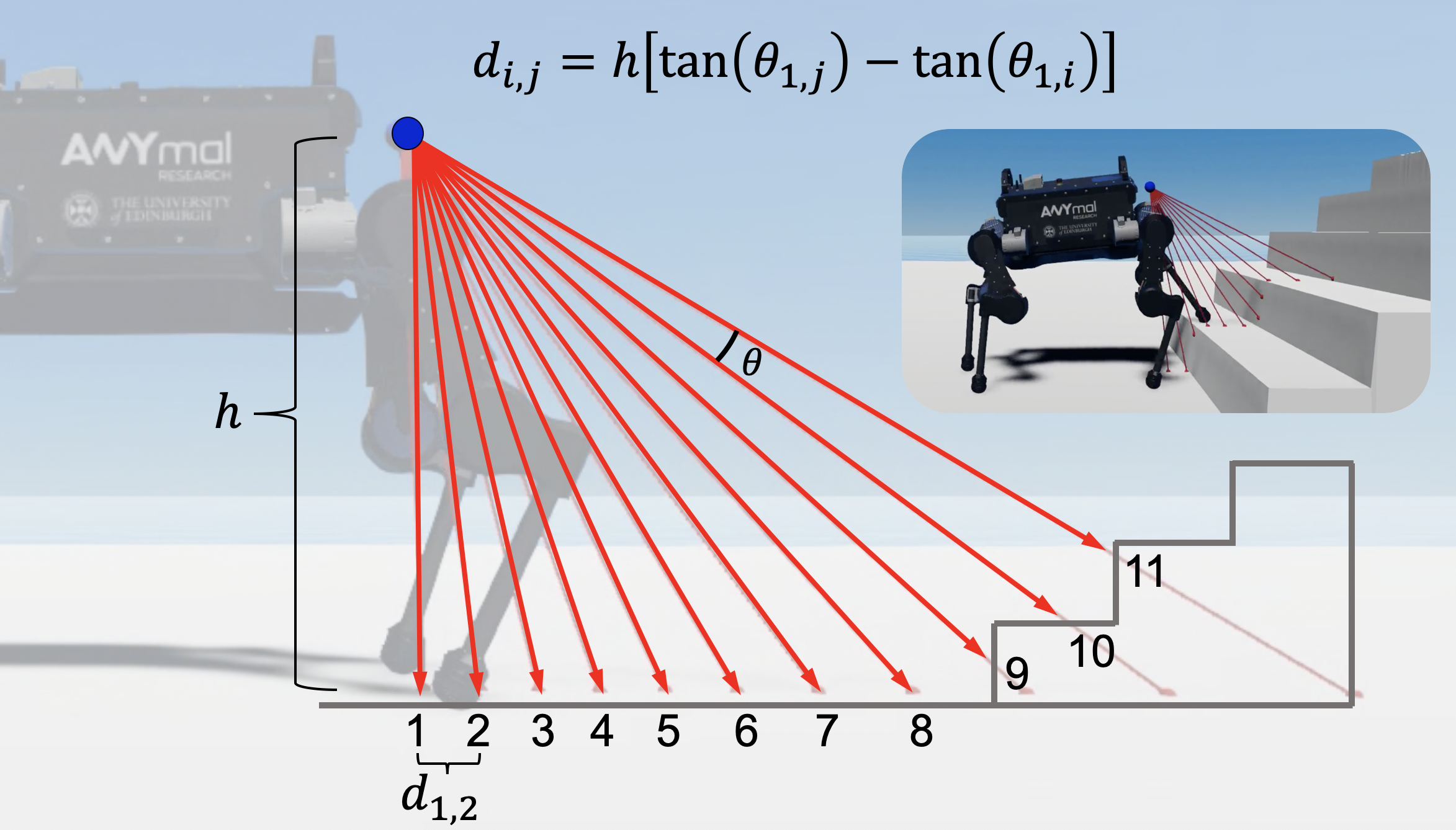}  
\caption{Design guide to determine the resolution of sparse observations for perceiving steppable terrain irregularities.
}
\label{fig:ray_res}
\vspace{-4mm}
\end{figure}

With regards to our choice of exteroceptive sensing, we are motivated by the capability to leverage small and affordable RGB-D and Lidar sensors to facilitate future deployment on real robot platforms. For this reason, we choose to replicate the vertical field-of-view of a popular commercial-off-the-shelf RGB-D sensor, the Intel RealSense D435, as shown in Figure \ref{fig:ray_res}. The number of rays used as an input for the perceptual locomotion policy was designed to provide sufficient spatial resolution and to perceive relevant features from the terrains selected, e.g., the presence of a step or stairs. 

We propose the following design guide to determine the number of depth observations: during locomotion, the size of the smallest segment between rays immediately ahead of the robot shall be smaller than its effective foot size, in order for the policy to be able to perceive surfaces where the feet can step on. The depth sensing device is oriented such that the edge of the field of view enables depth sensing in the downwards direction, therefore, the robot can perceive terrain features immediately in front and within its workspace.


Using Figure \ref{fig:ray_res}, on flat ground, the inter-ray distance is given by $d_{i,j} = h ( \tan{\theta_{1,j}} - \tan{\theta_{1, i}})$, where $\theta_{1, i}$ is the angle between rays 1 and $i$. In our case, the smallest inter-ray distance is found between rays 1 and 2, which for $N$ rays is given by $d_{1,2} = h ( \tan{\theta_{1,2}} -0 )= h \tan(\frac{60 
^{\circ}}{N-1})$. Given the dimensions of our robot, we have $h \approx 50$ cm and an effective foot size of $d_{\text{foot}} =$ 5.5 cm. Therefore, $N=11$ is the smallest number of rays that yields $d_{1,2} < d_{\text{foot}}$ cm. Whilst being a sparse observation of the environment ahead of the robot, we provide this design guide as a means to determine a sufficiently fine resolution to capture the presence of the smallest obstacle the robot could possibly step on within immediate proximity.

\subsection{Learning-based Control Architecture}
\label{sec:principle:ccc}

\begin{figure}
\centering
\includegraphics[width=80mm]{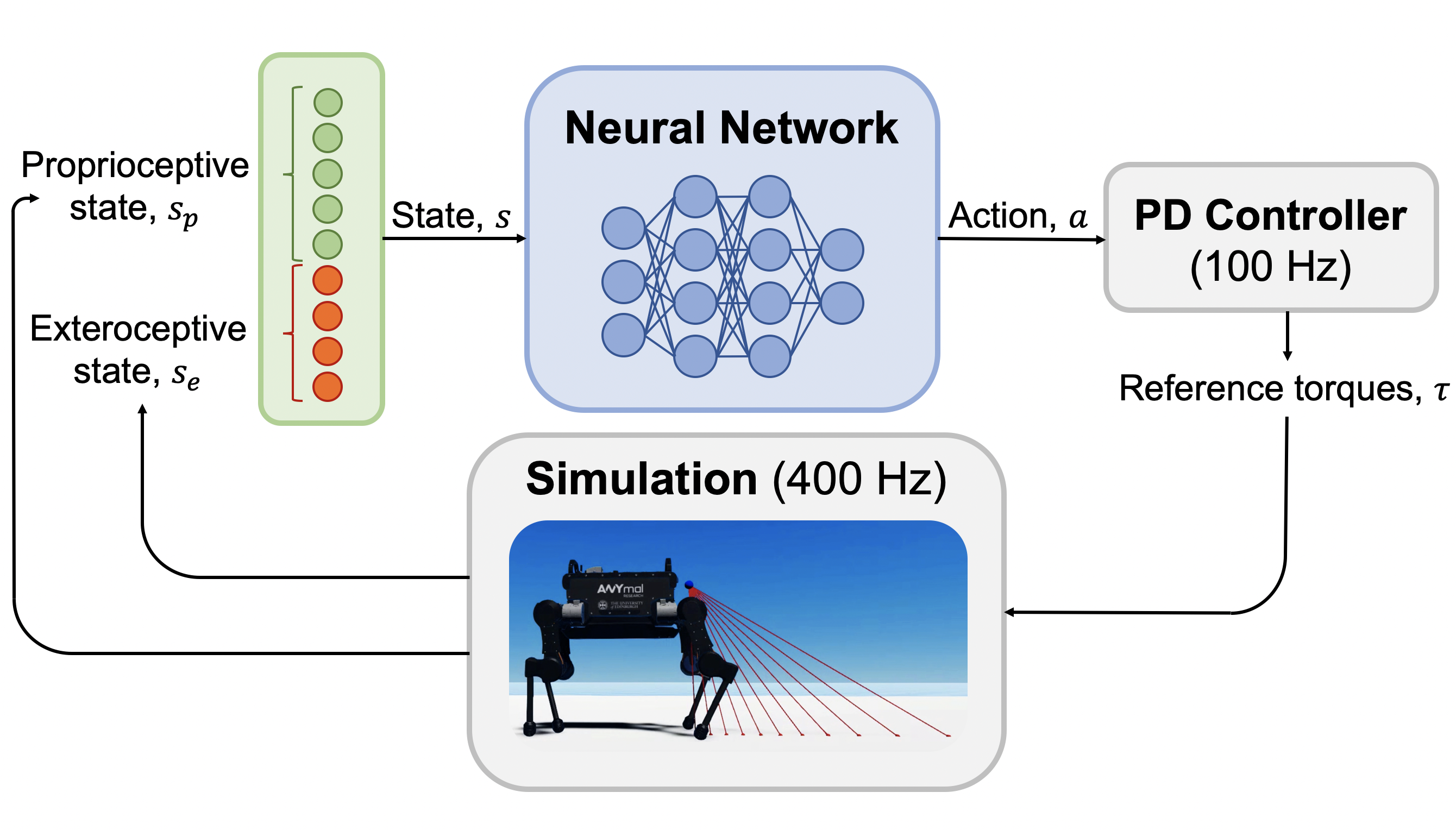} 
\caption{Control architecture used to train our neural network policy for perceptual locomotion.}
\label{fig:control_architecture}
\vspace{-3mm}
\end{figure}

The control architecture of our learning-based policy is depicted in Figure \ref{fig:control_architecture}, which is identical during training and testing. The neural network policy receives a state observation $s$ that consists of the concatenation of two vectors, a proprioceptive state vector $s_p$ and an exteroceptive state vector $s_e$. 

The proprioceptive state includes: body height in global coordinates $d_z$, trigonometric terms from the body rotation matrix encoding roll $\phi$ and pitch $\theta$, yaw angular error $\bar{\psi}$, body horizontal velocity error $(\bar{v_x},\bar{v_y})$, joint angles $q$, body linear velocity $(v_x, v_y, v_z)$, body angular velocity $(\dot{\phi}, \dot{\theta}, \dot{\psi})$, and joint angular velocities $\dot{q}$. Note that $\bar{\psi}$ and $(\bar{v_x},\bar{v_y})$ enable omnidirectional base control. The exteroceptive states $s_e$ are the distances measured from the 11 rays in our setup, where the measurement provides distances to physical obstacle, and is clipped between 0.1 m and 8 m which matches the specifications of commercially available RGB-D cameras. 

Fully-connected feedforward neural networks with hidden layer dimensions as in Table \ref{tab:ppo} are used for policy and value networks in an actor-critic framework. The policy network outputs action $a$, which represents the desired joint positions. These are fed into a PD controller running at 100 Hz to obtain reference joint torques, which are then used by the physics engine \cite{raisim} to time step the simulation at 400 Hz. The corresponding state measurements are then fed back into the policy as a closed-loop. In line with recent works \cite{MELA, RMA}, we design our policy to learn joint angles instead of joint torques as this avoids learning low-level joint dynamics, which are often poorly simulated, with the expectation to improve simulation-to-reality transfer of our method in the future.

\begin{table}
    \small
    \centering
    \caption{Reward parameters in the training environment.}
    \label{tab:parameters}
    \scalebox{0.95}{\begin{tabular}{|l|l|l|}
        \hline 
        $c_{\boldsymbol{\tau}} = -0.0001$            & $c_{\boldsymbol{\dot{q}}} = -0.0001$ & $c_{\phi, \theta} = -0.05$   \\
        \hline 
        $c_{\boldsymbol{v}} = 1$         & $c_{\psi} = -5$                      & $\bar{\psi}_{\text{clip}} = 0.3$  \\
        \hline 
    \end{tabular}}
    \label{tab:reward}
    \vspace{-3mm}
\end{table}

\subsection{Learning Locomotion via Reinforcement Learning}
\label{sec:principle:bbb}

\subsubsection{Reinforcement Learning}
\label{sec:principle:bbb:a}

We formulate the control problem as a Markov Decision Process (MDP), a mathematical formalism for modelling sequential decision processes, formally defined by a 4-tuple $\langle \mathcal{S}, \mathcal{A}, P(s_{t+1} \vert s_{t}, a_{t}), R \rangle$, where $ \mathcal{S}$ is the set of possible states, $ \mathcal{A}$ is the set of possible actions, $ P(s_{t+1} \vert s_{t}, a_{t})$ is the transition probability given a state-action pair and $R$ is a reward provided by the environment when transitioning into a new state. The control policy is represented by a neural network $\pi_{\theta}$. Proximal Policy Optimization (PPO) \cite{PPO}, an on-policy actor-critic RL algorithm, is used to train the policy and the PPO update is given by
\begin{align}
    \theta_{k+1} = \arg \max_{\theta} \mathbb{E}_{s,a \sim \pi_{\theta_k}} \left[
    L(s,a,\theta_k, \theta)\right]
\end{align}
with a clipped loss function $L(s,a,\theta_k, \theta)$ that has a surrogate term and an entropy term.

\subsubsection{Reward Design}
\label{sec:principle:bbb:b}
There are two groups of reward terms, those that are used for the general design of a locomotion policy and those that specifically enable the control policy to use state error terms $\bar{v_x},\bar{v_y}, \bar{\psi}$,  as control inputs in the feedback loop, where each error term is defined as the difference between the reference and measurement state. The reward terms that are designed to enable smooth locomotion are a torque term $r_{\boldsymbol{\tau}}$ given by 
\begin{align}
    r_{\boldsymbol{\tau}} = c_{\boldsymbol{\tau}} \vert \vert \boldsymbol{\tau} \vert \vert^2, \; c_{\boldsymbol{\tau}} < 0,
\label{eq:r_tau}
\end{align}
and a joint velocity term $r_{\boldsymbol{\dot{q}}}$ given by
\begin{align}
    r_{\boldsymbol{\dot{q}}} = c_{\boldsymbol{\dot{q}}} \vert \vert \boldsymbol{\dot{q}} \vert \vert^2, \;  c_{\boldsymbol{\dot{q}}} < 0 ,
\label{eq:r_qdot}
\end{align}
to penalize excessive use of actuators, and a body orientation term $ r_{\phi, \theta}$ given by
\begin{align}
    r_{\phi, \theta} = c_{\phi, \theta} \frac{\vert \phi \vert + \vert \theta \vert}{\pi}, \; c_{\phi, \theta} < 0,
\label{eq:r_pt}
\end{align}
to encourage the body to remain horizontal. 

Moreover, the reward terms associated with the control inputs are $r_{\boldsymbol{v}}$ given by
\begin{align}
    r_{\boldsymbol{v}} = 1 - c_{\boldsymbol{v}} \vert \vert \frac{\boldsymbol{v_{\text{ref}}} - \boldsymbol{v}}{\vert \vert \boldsymbol{v_{\text{ref}}} \vert \vert } \vert \vert^2, \; c_{\boldsymbol{v}} > 0,
\label{eq:r_v}
\end{align}
to minimize velocity error for any given reference velocity, and $r_{\psi}$ given by
\begin{align}
    r_{\psi} = c_{\psi} \min (\vert \frac{\psi_{\text{ref}} - \psi}{\pi} \vert, \bar{\psi}_{\text{clip}}), \; c_{\psi} < 0,
\label{eq:r_psi}
\end{align}
to minimize heading error up to a clipping value $\bar{\psi}_{\text{clip}}$. The total reward at any time is defined as the sum of rewards. Reward parameters used for training are in Table \ref{tab:reward}.


\begin{figure}
\centering
\includegraphics[width=80mm]{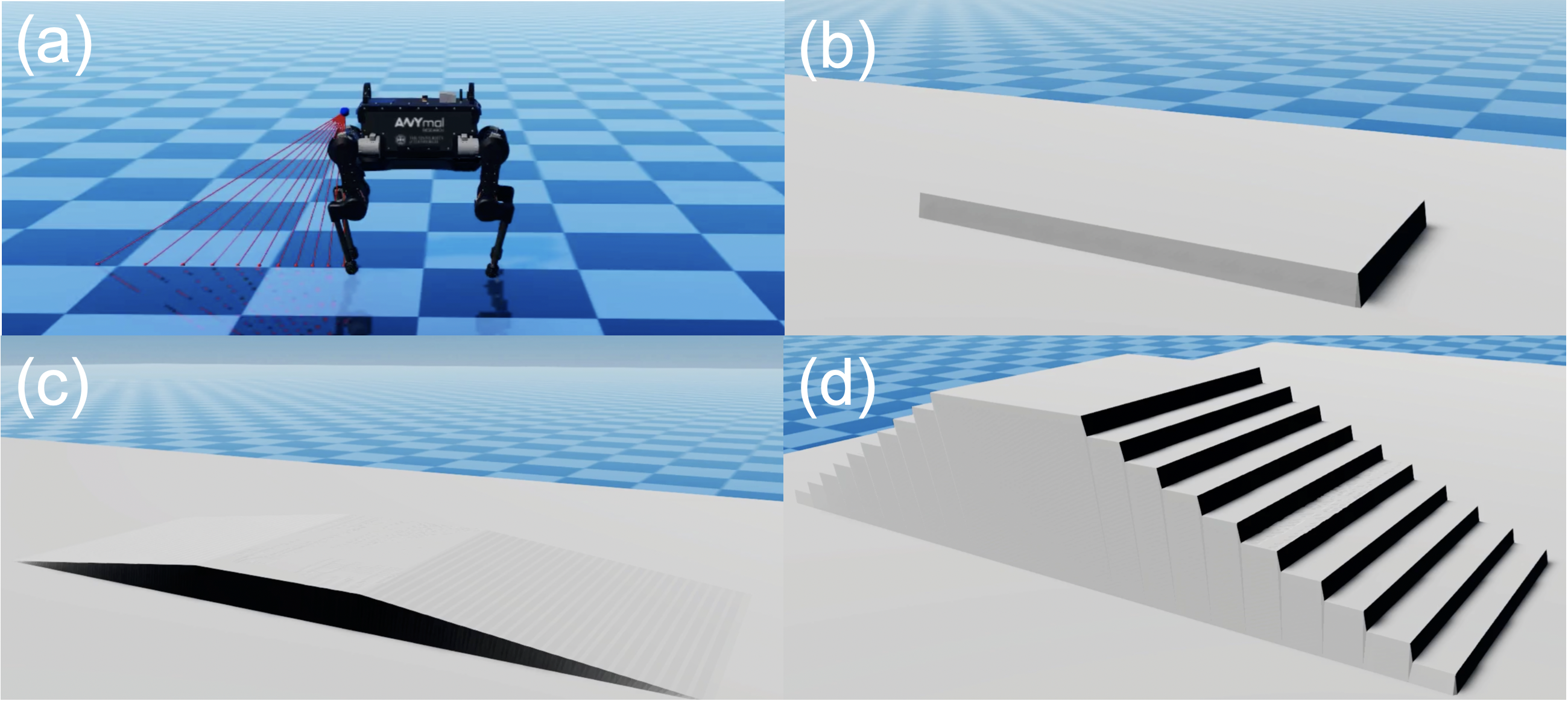}
\caption{Types of terrains used for training and testing: (a) flat ground, (b)  step, (c) ramp, (d) stairs.}
\label{fig:terrain_types}
\vspace{-4mm}
\end{figure}

\subsubsection{Curriculum Learning}
\label{sec:principle:bbb:c}


Whilst it may be possible to train a single policy directly by uniformly sampling the set of targeted tasks, due to inherent limitations with RL algorithms related to the exploration-exploitation dilemma and high-dimensional control problems, we designed a training curriculum to improve policy convergence for better sample efficiency. Two aspects are considered: first, training was conducted using 100 parallel environments, where a gradient step is taken using the averaged gradient across the environments; second, the training process followed a custom curriculum to easily allow the convergence of the control policy. At any point during training, the control task $\mathcal{T}_i$ for each of the parallel environments is sampled from a specified probability distribution $p(\mathcal{T})$, which is modified during the different stages of the training curriculum. Training a successful, convergent policy on the final stage of the curriculum \textit{ab initio} was not possible within our training experiments, which suggests that the training curriculum is instrumental to achieve successful convergence of the control policy. 

The training curriculum consists of a warm-up stage and four main stages. During each training stage, the first policy to achieve a minimum policy noise of 0.2 is selected, where noise is calculated as the standard deviation of action outputs from the mean output during that stage. In the warm-up stage, a control policy was trained on a uniform distribution of forward reference velocities [0, 0.7] m/s, where all values below 0.15 m/s are clipped to zero, and lateral reference velocity and reference yaw are zero. The resulting policy is then taken through four stages which modify $p(\mathcal{T})$ and $v_{\text{ref}}$ by including tasks in terrains shown in Figure \ref{fig:terrain_types}:

\begin{itemize}
\item Stage $\alpha$: introduce step, ramp, and stairs in the environment distribution with $p(\mathcal{T})$ of 0.1, 0.1, 0.5 respectively and 0.3 for flat floor. For step, ramp, and stairs, forward reference velocity is uniformly sampled from [0.15,0.7] m/s. For flat floor, reference velocity is sampled as in the warm-up stage. 
\item Stage $\beta$: same as previous stage, except that flat floor tasks include a lateral reference velocity component sampled from [-0.3, 0] m/s while reference yaw is zero, i.e., reference velocity can have forward and lateral left components. 
\item Stage $\gamma$: same as previous stage, except that flat floor tasks sample longitudinal velocity from [-0.7,0.7] m/s, i.e., a backward reference velocity component is introduced. 
\item Stage $\delta$: same as previous stage, except that flat floor tasks sample lateral velocity from [-0.3,0.3] m/s, i.e., a right lateral velocity component is introduced. 
\end{itemize}

The terrains in Figure \ref{fig:terrain_types} are designed based on their typical dimensions in environments inhabited by humans \cite{calistairs}. Ramp inclinations are between $5.7^{\circ}$-$11.3^{\circ}$, steps heights are between 10-20 cm and stairs step length is 30 cm.

\begin{figure}
\centering 
\subfloat[]{\includegraphics[clip, trim=0mm 0mm 0mm 0mm, width=78mm]{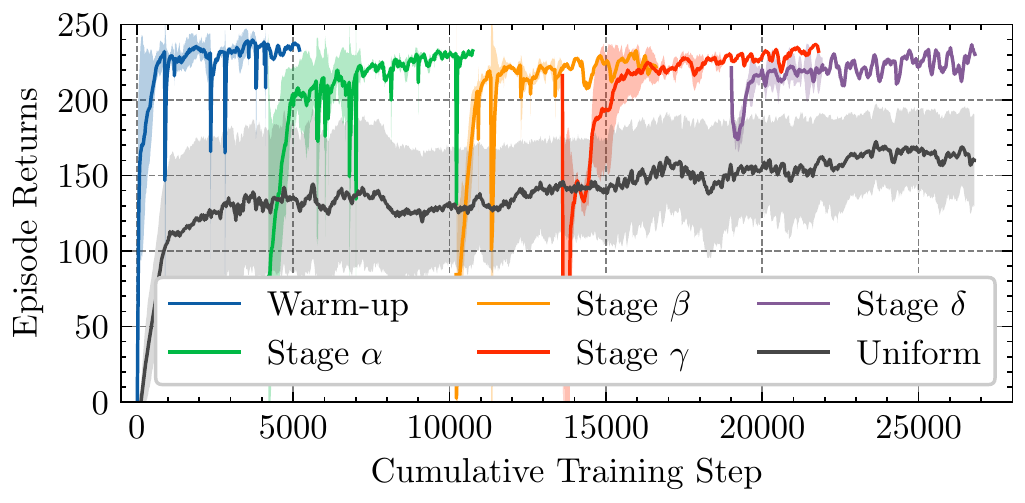}} \\[-5ex]
\hspace{0.08em}
\subfloat[]{\includegraphics[clip, trim=0mm 0mm 0mm 0mm, width=78mm]{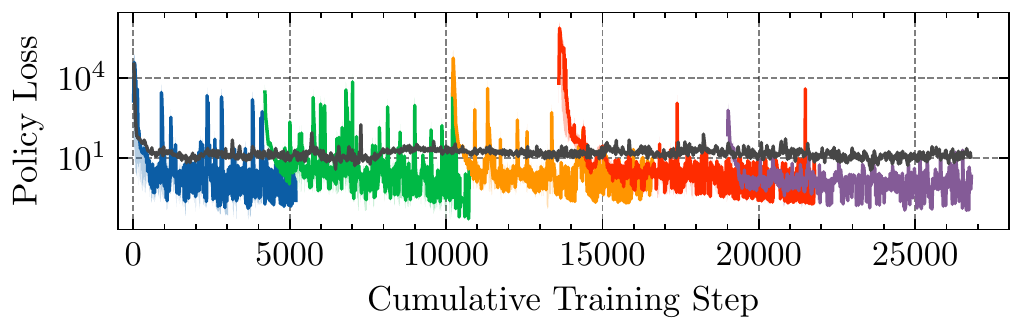}} \\[-5.5ex]
\hspace{0.08em}
\subfloat[]{\includegraphics[clip, trim=0mm 0mm 0mm 0mm, width=78mm]{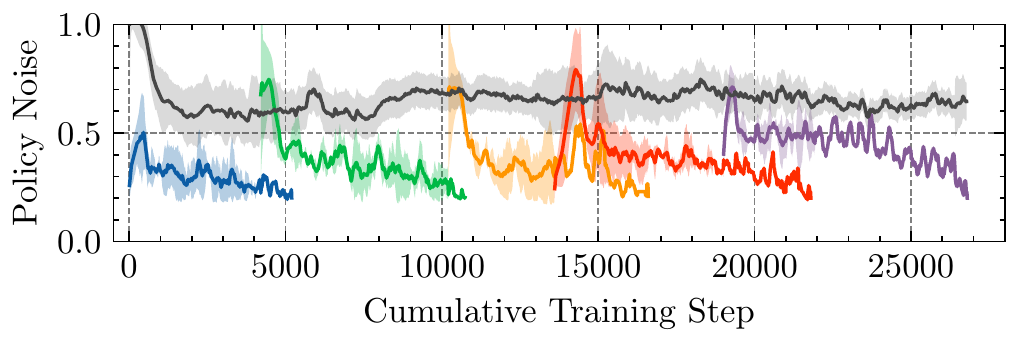}} \\[-3ex]
\caption{Episode returns, policy loss and noise during curriculum training from re-running our curriculum using 6 random seeds. Policy loss is the sum of surrogate loss and entropy loss. Results overlap between stages, as not all runs train each stage for the same number of steps. In contrast, \textit{Uniform} is trained from scratch on all tasks sampled uniformly, which yields lower performance. }
\label{fig:train}
\vspace{-2.5mm}
\end{figure}

\begin{table}
    \small
    \centering
    \caption{Parameters used during training.}
    \scalebox{0.95}{\begin{tabular}{|l|l|}
        \hline 
         Policy net: [128, 256, 128]             & Value net: [128, 256, 128] \\
        \hline 
         $\gamma = 0.996$       & $\lambda = 0.95$ \\
        \hline 
         $\epsilon = 0.2$            & $\alpha = 0.0002$ \\
        \hline 
         $c_1 = 0.5$     & $c_2 = 0.01$ \\
        \hline 
        $N_{\text{learning epochs}} = 4$     & $N_{\text{mini batches}} = 4$   \\
        \hline 
        Maximum gradient norm:  0.5     & Minimum policy noise: 0.2 \\
        \hline 
    \end{tabular}}
    \label{tab:ppo}
    \vspace{-4mm}
\end{table}

\section{Results of Perceptual Locomotion}
\label{sec:results}

This section presents the training process and evaluates the performance of the policy over seen and unseen terrains, and under the presence of exteroceptive noise and ablations.


\begin{figure*}[t]
\centering
\includegraphics[width=180mm]{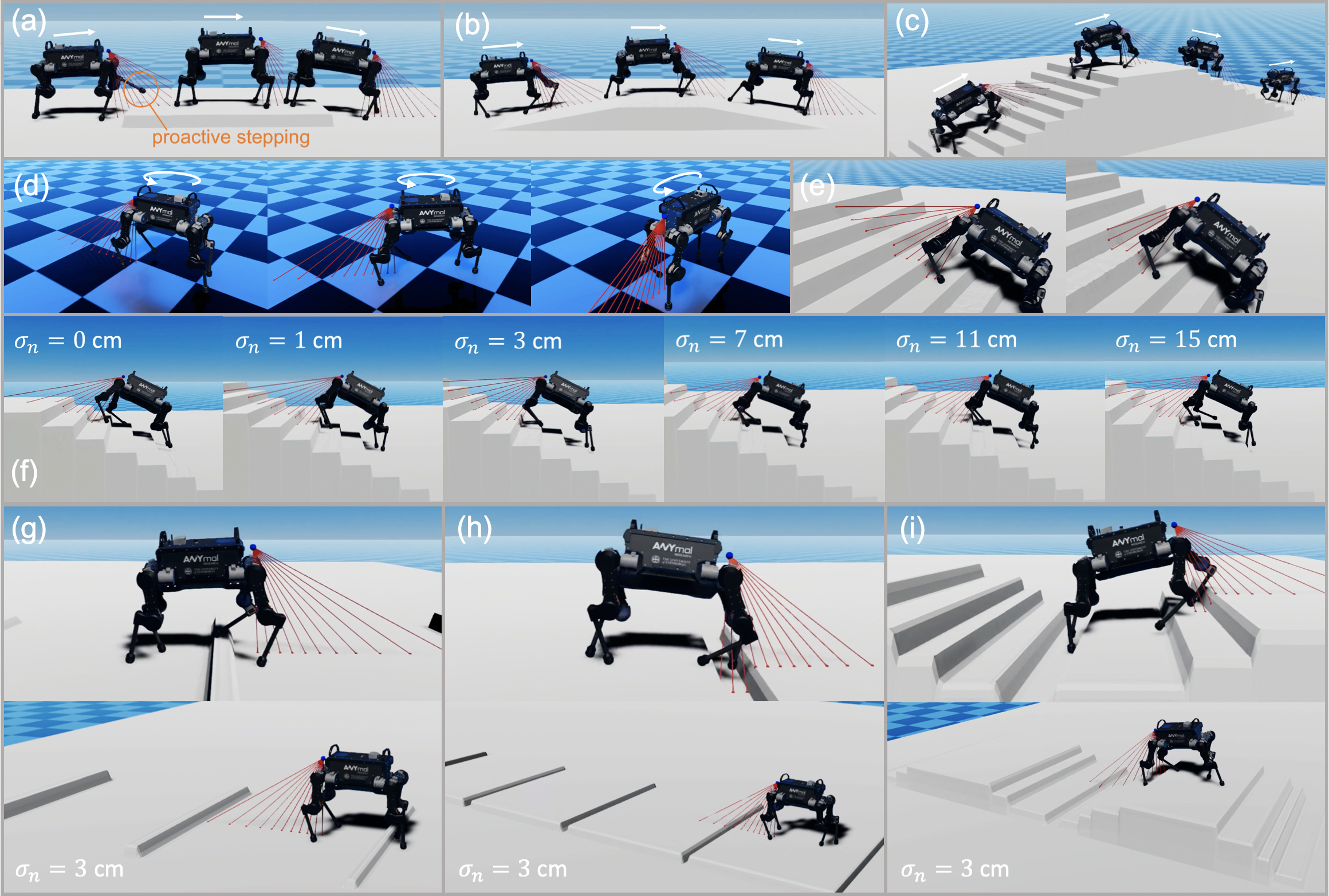}
\caption{ Perceptual locomotion policy tested on various seen and unseen tasks: 
(a)-(c) Tests on steps, ramps, and stairs; 
(d) Tests on flat ground with a turning command;
(e) Ablation tests of the exteroceptive observations;
(f) Tests on stairs using Gaussian noise with standard deviation $\sigma_n$ in exteroceptive observations;
(g)-(i) Tests in terrains not seen during training using noise, consisting of (g) barriers, (h) ditches, and (i) alternating stairs of varying step length.}
\label{fig:all_tests}
\vspace{-4mm}
\end{figure*}

\subsection{Training Process}
\label{sec:res:train}

The policy was trained using the parameters in Table \ref{tab:ppo}. The training curriculum enables the policy to learn the mapping required to control the robot in two groups of tasks: those that involve omnidirectional base movements on flat ground as well as those that involve perceptive locomotion over obstacles in forward direction. In contrast to our training curriculum, training from scratch by uniformly sampling all tasks from the final stage of the curriculum yields significantly worse performance. Using 6 random seeds, we provide in Figure \ref{fig:train} we show episode returns, policy loss and noise during the stages of curriculum training, as well as using uniform sampling across all tasks. In Figure \ref{fig:train}, a training step corresponds to updating the policy using $N_{\text{learning epochs}}$ gradient steps, each with $N_{\text{mini batches}}$ trajectories.

Results in Figure \ref{fig:train} show how the performance of the policies trained using our curriculum is significantly better than that of the uniform sampling runs. Policies trained with uniform sampling perform well on flat ground, but often trip or fail when stepping over obstacles, particularly in the stairs terrain, yielding the lower episode returns. Training with uniform sampling was stopped after the same number of steps in which our curriculum reaches our stop condition on policy noise. Considering the increasing trend of episode returns, it is possible that the uniform sampling approach could match performance of our curriculum, however this could happen after what we consider is prohibitively long time, especially compared to our training curriculum. In this sense, we argue that our training curriculum successfully aids policy convergence and improves sample efficiency to achieve top performance across our tasks. 

Using a desktop machine (8-core Intel i9 CPU, a 8GB GeForce RTX 2080 GPU), across all of our training runs, the average wall-clock training time for the warm-up training stage was 3.7 hours, stage $\alpha$ was 5.2 hours, stage $\beta$ was 5.3 hours, stage $\gamma$ was 6.3 hours, and stage $\delta$ was 6.5 hours. Due to the parallelized training environment, the real-time factor of the simulation during training was 300x, i.e. the policy was trained for an average total simulated time of 11.2 months. Most significantly, our preliminary training experiments showed that a blind locomotion policy which only uses the proprioceptive state $s_p$ as policy observation could be trained on the warm-up stage in 0.6 hours, which implies that the addition of the chosen exteroceptive state $s_e$ to train a perceptive locomotion policy does not increase training time prohibitively.

\begin{figure}
\centering 
\subfloat[]{\includegraphics[clip, trim=0mm 0mm 0mm 0mm, width=0.42\textwidth]{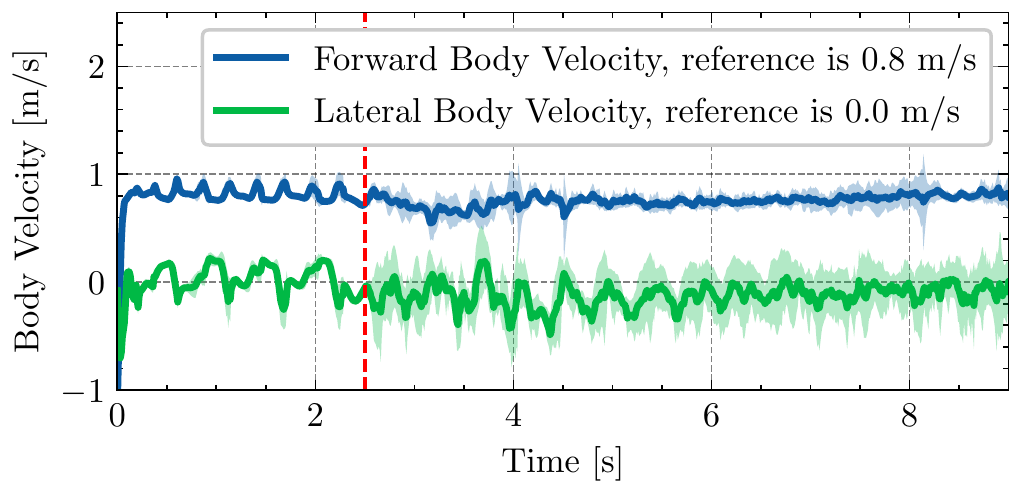}} \\[-5.5ex]
\hspace{0.08em}
\subfloat[]{\includegraphics[clip, trim=0mm 0mm 0mm 0mm, width=0.42\textwidth]{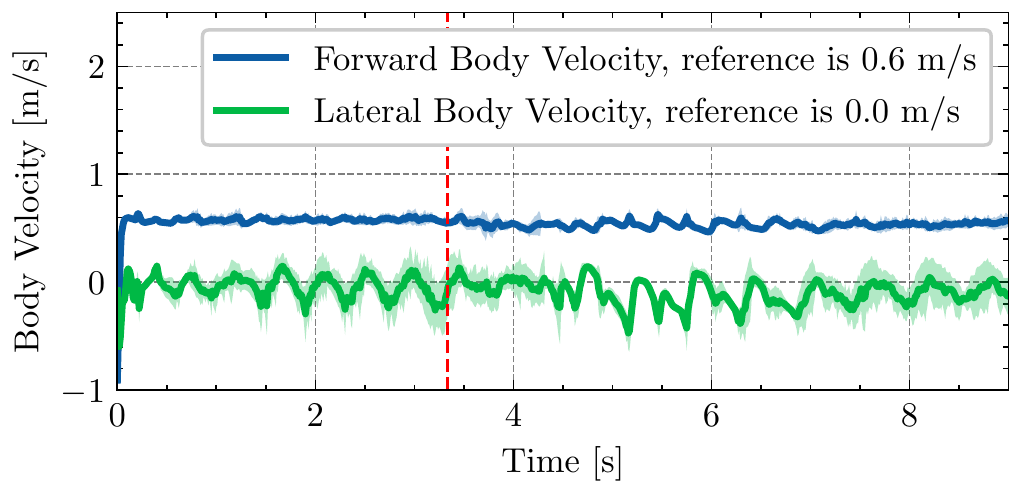}} \\[-5.5ex]
\hspace{0.08em}
\subfloat[]{\includegraphics[clip, trim=0mm 0mm 0mm 0mm, width=0.42\textwidth]{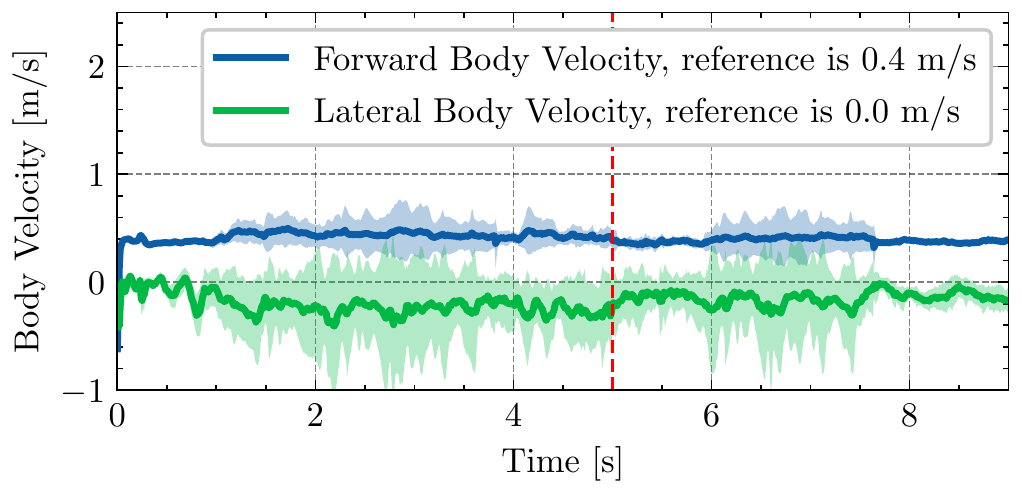}} \\[-5.5ex]
\hspace{0.08em}
\subfloat[]{\includegraphics[clip, trim=0mm 0mm 0mm 0mm, width=0.42\textwidth]{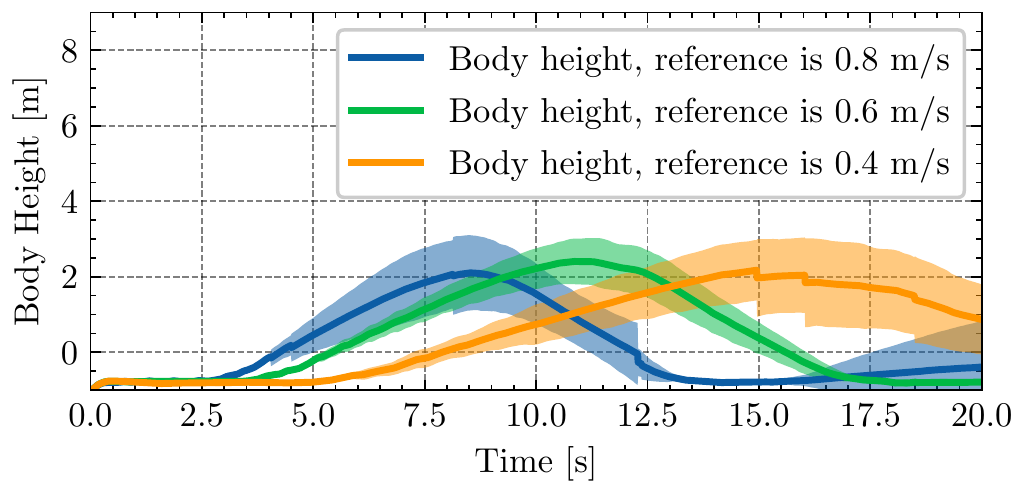}} \\[-3.5ex]
\caption{Body velocities and height measured during three sets of experiments on stairs. The dotted line on the velocity plots indicates the time at which the fore legs reach the first step of the stairs.}
\label{fig:stairs_diff_vels}
\vspace{-4mm}
\end{figure}

\begin{figure}
\centering 
\subfloat[]{\includegraphics[clip, trim=0mm 0mm 0mm 0mm, width=0.42\textwidth]{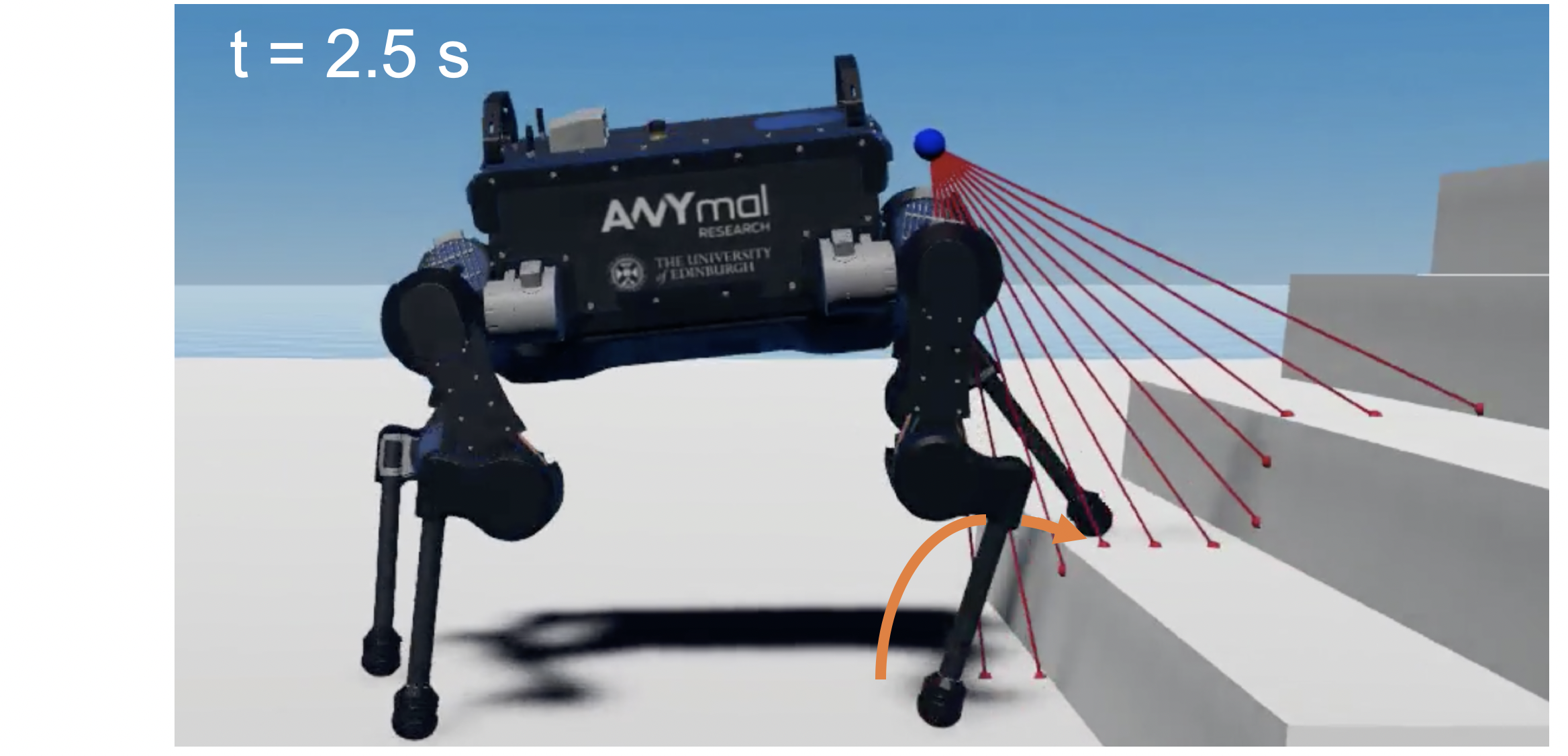}} \\[-5ex]
\hspace{0.08em}
\subfloat[]{\includegraphics[clip, trim=0mm 6mm 0mm 0mm, width=0.42\textwidth]{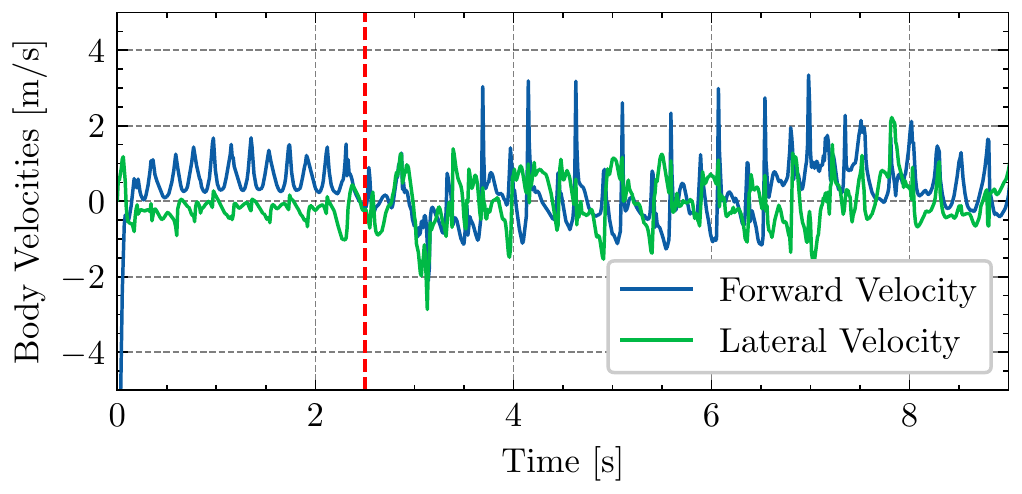}} \\[-5ex]
\hspace{0.08em}
\subfloat[]{\includegraphics[clip, trim=0mm 6mm 0mm 0mm, width=0.42\textwidth]{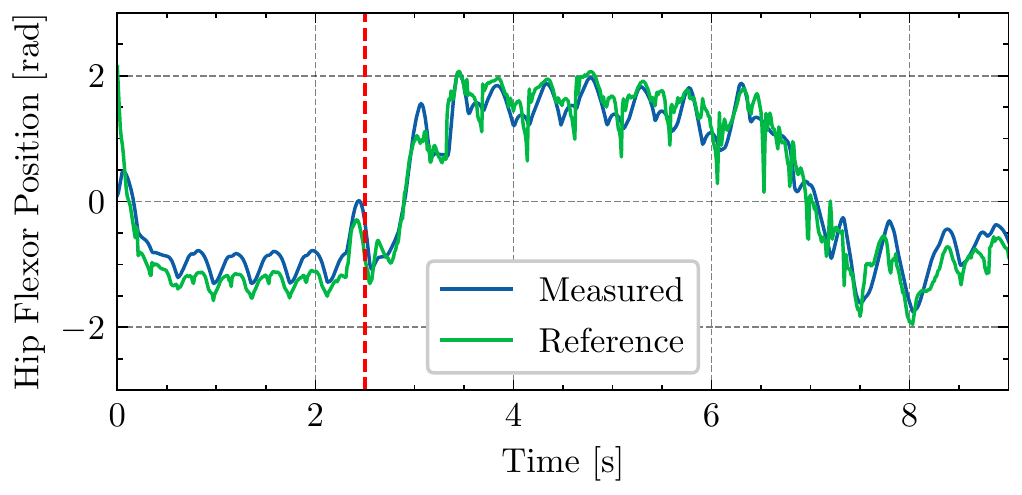}} \\[-5.5ex]
\hspace{0.08em}
\subfloat[]{\includegraphics[clip, trim=0mm 0mm 0mm 0mm, width=0.42\textwidth]{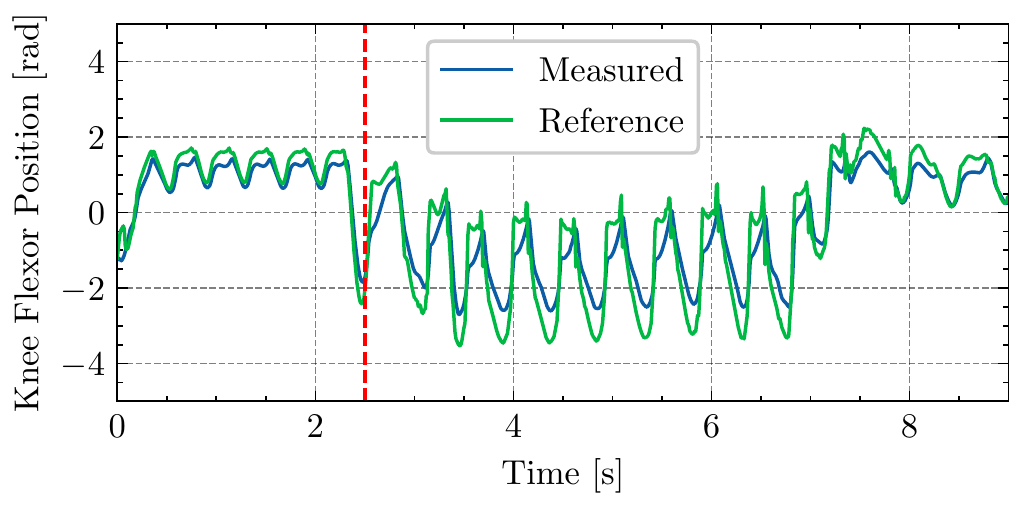}} \\[-3.5ex]
\caption{Stairs climbing depicted at $t=2.5$ seconds. Body velocities, hip and knee flexor joint trajectories are feature modification of the oscillatory motions during stairs climbing as well as increased velocity fluctuations.}
\label{fig:stairs_climbing_1}
\vspace{-5mm}
\end{figure}

\subsection{Forward Walking on Stairs}
\label{sec:res:terr}

The trained policy was extensively tested on the stairs as those in Figure \ref{fig:all_tests}(c), as it presents the highest difficulty for perceptual locomotion. Figure \ref{fig:stairs_diff_vels} shows body velocities and height measured during 3 sets of experiments with step height uniformly sampled between 10-20 cm, with reference forward velocities of $0.4$, $0.6$ and $0.8$ m/s. A total of 60 experiments were run per reference velocity. The velocity data in Figure \ref{fig:stairs_diff_vels} show how the standard deviation of the measured forward and lateral velocity increases during stairs climbing when compared to locomotion over flat ground before reaching the stairs. Lateral velocity presents the highest increase in standard deviation upon reaching the stairs, particularly for forward reference velocities of 0.4 m/s and 0.8 m/s, which can be explained by the lateral rocking motions require to climb the stairs at different reference forward velocities. Upon visual examination of the gaits, we believe that forward velocity of 0.4 m/s performs worse because it requires the greatest lateral rocking to maintain a stable gait when climbing such large stairs, which agrees with Table \ref{tab:success_rates} reporting the largest mean $\bar{v_y}$ error on stairs across forward velocities tested. 
 

\begin{table*}
\small
\centering
\caption{Summary of performance over different terrains from 60 repeated tests per task. Success rate is included, as well as mean squared error, mean and standard deviation for the error terms relating to the three control variables of the learned control policy, namely $\bar{v_x}$, $\bar{v_y}$, $\bar{\psi}$.}
\scalebox{0.92}{\begin{tabular}{|c|c|c|c|c|}
\hline
Task & \textbf{Success Rate} & $\bar{v_x}$ [m/s]  & $\bar{v_y}$ [m/s]  & $\bar{\psi}$ [deg]   \\ \thickhline
Stairs, $v_x = 0.8, v_y = 0$  & 93.3\% & 0.03, 0.02 $\pm$ 0.14  & 0.08, 0.07 $\pm$ 0.27 & 1.80, 0.26 $\pm$ 1.32\\ \hline
Stairs, $v_x = 0.6, v_y = 0$  & 100\% & 0.02, 0.01 $\pm$ 0.15  & 0.17, 0.15 $\pm$ 0.39 & 1.73, 0.15 $\pm$ 1.31\\ \hline
Stairs, $v_x = 0.4, v_y = 0$  & 91.7\% & 0.02, -0.01 $\pm$ 0.13  & 0.14, 0.18 $\pm$ 0.32 & 1.31, -0.04 $\pm$ 1.15\\ \hline
Ramp, $v_x = 0.8, v_y = 0$  & 100\% & 0.04, 0.14 $\pm$ 0.13  & 0.05, -0.05 $\pm$ 0.22 & 1.08, 0.09 $\pm$ 1.04\\ \hline
Ramp, $v_x = 0.6, v_y = 0$  & 100\% & 0.07, 0.22 $\pm$ 0.16  & 0.15, -0.04 $\pm$ 0.39 & 0.39, -0.16 $\pm$ 0.60\\ \hline
Ramp, $v_x = 0.4, v_y = 0$  & 100\% & 0.05, 0.20 $\pm$ 0.08  & 0.08, -0.02 $\pm$ 0.29 & 0.30, -0.19 $\pm$ 0.51\\ \hline
Step, $v_x = 0.8, v_y = 0$  & 100\% & 0.02, -0.02 $\pm$ 0.13  & 0.05, 0.00 $\pm$ 0.23 & 1.42, 0.09 $\pm$ 1.19\\ \hline
Step, $v_x = 0.6, v_y = 0$  & 100\% & 0.01, -0.04 $\pm$ 0.18  & 0.03, 0.04 $\pm$ 0.18 & 0.92, -0.07 $\pm$ 0.96\\ \hline
Step, $v_x = 0.4, v_y = 0$  & 100\% & 0.05, -0.02 $\pm$ 0.22  & 0.09, 0.11 $\pm$ 0.32 & 1.34, -0.16 $\pm$ 1.14\\ \hline
Flat ground, $v_x = 0.7, v_y = 0$  & 100\% & 0.01, -0.01 $\pm$ 0.09  & 0.04, -0.01 $\pm$ 0.19 & 0.69, -0.12 $\pm$ 0.82\\ \hline
Flat ground, $v_x = -0.7, v_y = 0$  & 100\% & 0.01, 0.03 $\pm$ 0.08  & 0.03, 0.01 $\pm$ 0.10 & 0.30, -0.06 $\pm$ 0.55\\ \hline
Flat ground, $v_x = 0.5, v_y = 0$  & 100\% & 0.00, -0.03 $\pm$ 0.06  & 0.03, 0.11 $\pm$ 0.14 & 0.42, -0.15 $\pm$ 0.63\\ \hline
Flat ground, $v_x = -0.5, v_y = 0$  & 100\% & 0.01, 0.07 $\pm$ 0.05  & 0.01, 0.04 $\pm$ 0.09 & 0.27, -0.08 $\pm$ 0.51\\ \hline
Flat ground, $v_x = 0, v_y = 0.3$  & 100\% & 0.04, -0.03 $\pm$ 0.18  & 0.05, 0.02 $\pm$ 0.09 & 0.24, -0.09 $\pm$ 0.48\\ \hline
Flat ground, $v_x = 0, v_y = -0.3$  & 100\% & 0.03, 0.02 $\pm$ 0.16  & 0.03, 0.01 $\pm$ 0.10 & 0.19, -0.09 $\pm$ 0.42\\ \hline
Flat ground, $v_x = 0.3, v_y = 0.3$  & 100\% & 0.04, 0.03 $\pm$ 0.14  & 0.04, -0.03 $\pm$ 0.19 & 0.32, -0.02 $\pm$ 0.56\\ \hline
Flat ground, $v_x = -0.5, v_y = -0.3$  & 100\% & 0.05, 0.06 $\pm$ 0.22  & 0.03, 0.07 $\pm$ 0.14 & 0.33, -0.03 $\pm$ 0.57\\ \hline
\end{tabular}}
\label{tab:success_rates}
\vspace{-5mm}
\end{table*}

\begin{table*}
\small
\centering
\caption{Summary of performance on the stairs task with $v_x = 0.6, v_y = 0$, where metrics are obtained from 60 repetitions per task. Success rate per task is included, as well as the \textbf{ratio} of error terms with performance in the noiseless case for this task as shown in Table \ref{tab:success_rates}. Error terms are as in Table \ref{tab:success_rates}. Gaussian noise standard deviation is given by $\sigma_n$.}
\scalebox{0.95}{\begin{tabular}{|c|c|c|c|c|}
\hline
 Exteroceptive Noise $\sigma_n$ [m] & \textbf{Success Rate} & Ratio $\bar{v_x}$ [-]  & Ratio $\bar{v_y}$ [-]  & Ratio $\bar{\psi}$ [-]   \\ \thickhline
 0.01 & 100\% & 0.81, 0.85 $\pm$ 0.89  & 0.92, 1.11 $\pm$ 0.95 & 0.74, 0.96 $\pm$ 0.85\\ \hline
 0.03 & 100\% & 0.91, 0.94 $\pm$ 0.95  & 0.92, 0.86 $\pm$ 0.97 & 0.76, 0.91 $\pm$ 0.87\\ \hline
 0.07 & 100\% & 1.28, 1.63 $\pm$ 1.12  & 1.17, 1.36 $\pm$ 1.06 & 1.00, 0.93 $\pm$ 1.01\\ \hline
 0.11 & 73.33\% & 1.45, 1.55 $\pm$ 1.21  & 1.55, 1.60 $\pm$ 1.22 & 1.11, 1.16 $\pm$ 1.04\\ \hline
 0.15 & 51.67\% & 2.5, 1.67 $\pm$ 1.59  & 2.54, 1.59 $\pm$ 1.57 & 1.75, 1.11 $\pm$ 1.35\\ \hline

\end{tabular}}
\label{tab:success_rates_noise}
\vspace{-5mm}
\end{table*}

\begin{table}
    \centering
    \caption{Success rates in terrains not seen during training, tested with exteroceptive noise $\sigma_n = 3$ cm.}
\scalebox{0.95}{\begin{tabular}{|c|c|}
        \hline 
        Terrain            & \textbf{Success Rate}    \\
        \thickhline 
        Barriers        &       78.33$\%$            \\
        \hline
        Ditches        &       71.16$\%$            \\
        \hline
        Alternating stairs (full)        &       56.66$\%$            \\
        \hline
        Alternating stairs (first block)        &       96.66$\%$            \\
        \hline
    \end{tabular}}
    \label{tab:success_unseen}
    \vspace{-4mm}
\end{table}

\vspace{-2mm}
\subsection{Analysis of Perception-Action Reflex}
\label{sec:res:reflex}


The introduction of perceptual information in the state observation enables the control policy to anticipate the presence of obstacles and adapt the gait as needed to overcome the obstacle, as in Figures \ref{fig:all_tests}(a) and \ref{fig:stairs_climbing_1}: the responsive motion of the fore legs \textit{anticipate} and react to the presence of obstacles. Figure \ref{fig:stairs_climbing_1} presents the joint trajectories of fore leg joint positions during stairs climbing, and show how the perceptual state inputs corresponds to a perception-action reflex that allows the policy to step over the steps without colliding.

Further, particular tests were additionally performed to analyze the influence of the exteroceptive inputs in policy performance, where one or more rays were forcefully set to the minimum allowed threshold. These ablation tests, illustrated in Figure \ref{fig:all_tests}(e), show that when any one of the 8th-11th rays (as per the numeration in Figure \ref{fig:ray_res}) is ablated, the performance on various terrains remains within $5\%$ difference in success rate of that in Table \ref{tab:success_rates}, i.e., tests without any ablation. 

However, when all 8th-11th rays (as per Figure \ref{fig:ray_res}) were ablated, the success rate on stairs dropped to $40-55\%$, depending on the reference forward velocity, with failure typically observed while approaching or during stairs descent. Whenever any one or multiple rays from 1-7 as per Figure \ref{fig:ray_res} were ablated, the policy was unable to succeed on any terrain. This is to be expected as the policy does not have any recurrent connections within the neural network, i.e., the policy is memory-less and thus heavily relies on exteroceptive observations in close proximity to the fore legs. In conclusion, our analysis highlights the level of importance of perceptual inputs which trigger the perception-action reflex and enable successful locomotion over the terrains tested.

\vspace{-2mm}
\subsection{Statistical Performance on Seen Terrains}
Results of the policy performance on tasks involving terrains seen during training are in Table \ref{tab:success_rates}, such terrains are shown in Figure \ref{fig:all_tests}(a)-(c). For tasks on step, ramp, and stairs, tests are presented with forward reference velocities of 0.4, 0.6, and 0.8 m/s. Terrain heights were uniformly sampled from values ranging between 10-20 cm for step height of the step and stairs terrains, and slope angles of $5^{\circ}$-$11^{\circ}$ for the ramp terrain. For tasks on flat ground, forward reference velocities of $\pm$0.7 and $\pm$0.5 m/s are tested, as well as lateral reference velocities of $\pm$0.3 m/s, and two tasks with omnidirectional base commands that require diagonal walking with forward and lateral velocities (0.3, 0.3) and (-0.5, -0.3) m/s. 

The results shown in Table \ref{tab:success_rates} show the high success rate of the trained policy, where all tasks present 100$\%$ success except for two tasks on stairs, which nevertheless present success above 90$\%$. The training curriculum also enables the policy to generalize to perform turning while walking as shown in Figure \ref{fig:all_tests}(d), achieved by constantly adjusting $\psi_{\text{ref}}$ such that $\vert  \psi_{\text{ref}} - \psi \vert = 0.3$, whereas in training $\psi_{\text{ref}} = \psi_{\text{initial}}$. 

Moreover, we evaluated the performance of the trained policy for the stairs task with five levels of directly added Gaussian noise with standard deviation $\sigma_n$ in the exteroceptive measurements, as shown in Table \ref{tab:success_rates_noise} and Figure \ref{fig:all_tests}(f). The success rate of the policy is not affected while $\sigma_n$ is within 1-7 cm, which well covers the level of noise exhibited by commercially available hardware for the depth values we consider. Our evaluation found that performance degrades progressively as $\sigma_n$ continues to increase to 11-15 cm. While the policy was trained in a noiseless setting, the above tests show that performance is unaffected for low values of added noise within 7 cm. We hypothesize this may be due to the oscillatory motion the body exhibits during training which in itself yields small fluctuations in the depth measurements, although this hypothesis remains untested and could be incorrect as these fluctuations are correlated to one another and to the robot joint configuration. 



\vspace{-2mm}
\subsection{Statistical Performance on Unseen Terrains}
We test the policy on various terrains not seen during training to evaluate its generalizability and determine if the proposed training curriculum is well designed to enable one single policy to traverse different terrains. To do so, we test the policy in three unseen terrains as shown in Figure \ref{fig:all_tests}(g)-(i): barriers (g), ditches (h), and alternating stairs (i). We perform these tests at reference velocity $v_x = 0.6, v_y = 0$, and with $\sigma_n = 3$ cm to further demonstrate policy robustness. The barriers consist of square sections of 10, 15, and 20 cm. The ditches are platforms raised above ground with inter-platform separation of 5, 10, 15, 20, 30, 40 cm, and height of 15 cm. The alternating stairs consist of two blocks of stairs with steps lengths varying between 20 and 55 cm, whereas our policy was only trained on step length 30 cm. Success rates for the tests in these unseen terrains with exteroceptive noise are in Table \ref{tab:success_unseen}, obtained from 60 trials. 

The policy generalizes well to barriers and ditches. In alternating stairs, we found that the most frequent failures occur during the transition between two blocks of stairs, while the hind legs are still on the last step of the first block. Since the policy is memory-less and the forward-facing visual observations only see the front, the hind legs have limited ability to overcome the gap between stairs, resulting in a lower success rate of 56.66$\%$. The first block of stairs was traversed with success rate 96.66$\%$, demonstrating adaptability to stairs of varying step lengths. 



\section{Conclusion and Future Work}
\label{sec:conclusion}


This work investigates a learning scheme for perceptual locomotion using only sparse visual feedback. Our policy is able to learn the state-action mapping with exteroceptive observations and perform successfully on steps, ramps and stairs. The latter terrains are particularly the cases where blind locomotion policies cannot achieve the ascent or descent of high stairs at a 0.8 m/s velocity command. Further, our policy is robust to exteroceptive noise and ablations, and generalizes well to multiple unseen terrains featuring barriers, ditches, and alternating stairs.

The results suggest a potential of this method to be used for producing more affordable robotic solutions, which can be equipped with simple, low-cost, and low-weight vision systems while still being able to accomplish designated industrial applications. As commercially-available quadruped robots have reduced cost significantly (less than \$2,000 per unit), to achieve cost-effective perceptual locomotion can generate a big momentum to populate the use of legged machines in a wider range of industrial and human-centered applications. 

Within the scope of this work, our particular design is to investigate whether perceptual locomotion can be learned by using sparse visual inputs, so the exteroceptive observations are kept at a very low dimension, as we are exploring computationally efficient and functionally effective solutions. However, this simplicity trades off the completeness of information: the policy only sees the height in-between two legs; when terrain heights are asymmetric between the left/right leg, sparse rays do not scan these details; though the robot demonstrated good performance on steps and stair-like surfaces, such low scanning resolution is inevitably limited on more complex surfaces or smaller obstacles. Hence, for future work, it will be interesting to exploit the use of attention mechanisms for visual perception where more detailed scanning will be triggered only when necessary, and otherwise sparse observations are used for lower computational power and longer mission time.


\bibliography{root}
\balance
\bibliographystyle{IEEEtran}

\end{document}